%% file: main.tex
\documentclass[conference]{IEEEtran}
\IEEEoverridecommandlockouts
\input{preamble}

\def\BibTeX{{\rm B\kern-.05em{\sc i\kern-.025em b}\kern-.08em
    T\kern-.1667em\lower.7ex\hbox{E}\kern-.125emX}}
\begin{document}

\title{Trusted Confidence Bounds for Learning Enabled Cyber-Physical Systems}

\author{\IEEEauthorblockN{Dimitrios Boursinos}
\IEEEauthorblockA{\textit{Institute for Software Integrated Systems} \\
\textit{Vanderbilt University}\\
Nashville TN, USA \\
dimitrios.boursinos@vanderbilt.edu}
\and
\IEEEauthorblockN{Xenofon Koutsoukos}
\IEEEauthorblockA{\textit{Institute for Software Integrated Systems} \\
\textit{Vanderbilt University}\\
Nashville TN, USA \\
xenofon.koutsoukos@vanderbilt.edu}
}

\maketitle

\input{abstract.tex}
\input{keywords.tex}
\input{introduction.tex}
\input{problem.tex}
\input{triplet.tex}
\input{background.tex}
\input{evaluation.tex}
\input{conclusion.tex}
\input{aknowledgement}

\end{document}

%% file: preamble.tex
\usepackage{cite}
\usepackage{amsmath,amssymb,amsfonts}
\usepackage{graphicx}
\usepackage{textcomp}
\usepackage{xcolor}
\usepackage{bm}
\usepackage{tabularx}
\usepackage{multirow}
\usepackage{tikz}
\usetikzlibrary{positioning, calc, fit, shapes, arrows}
\usepackage{subcaption}
\usepackage [english]{babel}
\usepackage [autostyle, english = american]{csquotes}
\MakeOuterQuote{"}

\usepackage{algorithm}
\usepackage{algpseudocode}

%% file: abstract.tex

\begin{abstract}
Cyber-physical systems (CPS) can benefit by the use of learning enabled components (LECs) such as deep neural networks (DNNs) for perception and decision making tasks. However, DNNs are typically non-transparent making reasoning about their predictions very difficult, and hence their application to safety-critical systems is very challenging. LECs could be integrated easier into CPS if their predictions could be complemented with a confidence measure that quantifies how much we trust their output. The paper presents an approach for computing confidence bounds based on Inductive Conformal Prediction (ICP). We train a Triplet Network architecture to learn representations of the input data that can be used to estimate the similarity between test examples and examples in the training data set. Then, these representations are used to estimate the confidence of set predictions from a classifier that is based on the neural network architecture used in the triplet. The approach is evaluated using a robotic navigation benchmark and the results show that we can computed trusted confidence bounds efficiently in real-time. 
\end{abstract}

%% file: keywords.tex
\begin{IEEEkeywords}
Cyber-physical systems, deep neural networks, assurance monitoring, conformal prediction, triplet, robot navigation
\end{IEEEkeywords}

%% file: introduction.tex
\section{Introduction}
Machine learning components are being used by many cyber-physical system (CPS) applications because of their ability to handle dynamic and uncertain environments. Deep neural networks (DNNs), for example, are used for perception and decision making tasks in autonomous vehicles. Although such components offer many advantages for representing knowledge  in high-dimensional spaces and approximating complex functions, they introduce significant challenges when they are integrated into the system. 
Typical DNNs are non-transparent and it is not clear how to rationalize their predictions. Modern architectures are parameterized
using million of values which makes reasoning about their predictions very challenging.

Complementing the predictions of DNNs with a confidence measure can be very useful for improving the trustworthiness of such models and allow their application to safety critical systems. Several approaches have been proposed for confidence estimation. Neural networks for classification, in particular, typically provide probability-like outputs using a softmax layer. However, these probabilities are typically overconfident even for inputs coming from the same distribution as the training data~\cite{Guo:2017:CMN:3305381.3305518}. The reason is that the softmax probabilities are not well-calibrated meaning they do not provide a good estimate of the error rates. Several methods are proposed to compute well-calibrated confidence values. A class of methods aims to estimate scaling factors from the training data and use them to scale the softmax probabilities in order to compute well-calibrated confidence values and include temperature scaling~\cite{Guo:2017:CMN:3305381.3305518}, Platt scaling \cite{Platt99probabilisticoutputs}, and isotonic regression \cite{Zadrozny:2002:TCS:775047.775151}. Although such methods can compute well-calibrated confidence values, using them in CPS requires selecting an appropriate confidence bound which ensures a very small error rate and at the same time limiting the number of inputs for which a confidence prediction cannot be made.

The approach presented in this paper is based on conformal prediction
(CP)~\cite{balasubramanian2014conformal}. For classification, CP associates reliable confidence values with set predictions that may include multiple labels. The confidence measures are well-calibrated and can be computed in an online setting which is
suitable for CPS applications.
The online application of the approach is based on the inductive conformal prediction (ICP)
for computational efficiency~\cite{papadopoulos2008inductive}. ICP leverages a calibration data set that is used to compute the confidence values of new previously unseen inputs efficiently. The confidence values rely on nonconformity functions computed using techniques such as $k$-Nearest
Neighbors and Kernel Density Estimation. However, such approaches do not
scale for high-dimensional inputs.

In our previous work, we used the ICP framework for assurance
monitoring of CPS with machine learning components~\cite{boursinos2020assurance}.
In order to handle high-dimensional inputs in real-time, the approach computes
the nonconformity scores using the embedding representations
produced by trained DNN models in lower dimensional spaces than the input space. 
The main contribution of this paper is a significant improvement in computing confidence bounds by employing a triplet network architecture 
to learn representations of the input data that can be used to estimate the similarity between test examples and examples 
in the training data set. Then, these representations are used to estimate the confidence of set predictions from a classifier that is based on the neural network architecture used in the triplet. In order to achieve the desired confidence, the LEC may need to generate more than 
one possible prediction. Sets with multiple predictions can be useful especially when the alternatives are provided to a human operator. 
However, in this paper, we focus on autonomous 
CPS and compute the optimal confidence bounds required for the LEC to generate single predictions.

Another contribution of the paper is the evaluation of the approach using the SCITOS-G5 robotic navigation benchmark. 
Our results show that we can compute trusted confidence bounds efficiently in real-time.
We also compare the proposed approach with the approach presented in~\cite{boursinos2020assurance} which relies on applying ICP using 
representations learned by the model used to generate the predictions. The comparison shows for a chosen confidence bound, ICP using triplet produces fewer sets with multiple candidate labels, or equivalently there is a higher confidence bound that eliminate all the sets with multiple predictions.

The paper is organized a follows. Section~\ref{sec:problem} describes the problem. Section \ref{sec:triplet} reviews the Triplet Network architecture and the associated training algorithms. Section~\ref{sec:background} presents the methods for ICP using the Triplet.
We evaluate the performance of the proposed approach in Section~\ref{sec:evaluation} and we discuss the conclusions in Section~\ref{conclusion}. 

%% file: problem.tex
\section{Problem}
\label{sec:problem}
Learning-enabled components in CPS are used for taking decisions based on the state of the system and the environment. For a mobile robot, for example, an LEC can be used to navigate through a room while avoiding collisions with obstacles. A commonly used approach involves learning a model using training data and using the learned model for operation. We expect the system to perform better in scenarios similar to those used during the training phase. The problem we consider is the computation of a significance level along with each decision by the LEC. The significance level must be well-calibrated which means that it must be consistent with the expected error and ideally the expected error rate must be bounded. Moreover, CPS applications require minimizing the number of incorrect decisions while limiting the number of false alarms to enable efficient monitoring. 
Another requirement for CPS is that such monitoring must be performed in real-time often with limited computional resources.


%% file: triplet.tex
\section{Triplet Network}
\label{sec:triplet}
The ICP framework requires a way to estimate the similarity between the training data and a test input. 
The main idea in our approach is that we can do this efficiently by learning representations of the inputs
for which the Euclidean distance is a suitable measure of similarity. 
The triplet network is a Deep Neural Network (DNN) architecture that is trained to compute appropriate representations for metric or distance learning~\cite{hoffer2014deep}.

\begin{figure}[htb]
\centering
\input{images/triplet_architecture.tikz}
\caption{Triplet network structure \cite{hoffer2014deep}}
\label{fig:triplet_architecture}
\end{figure}
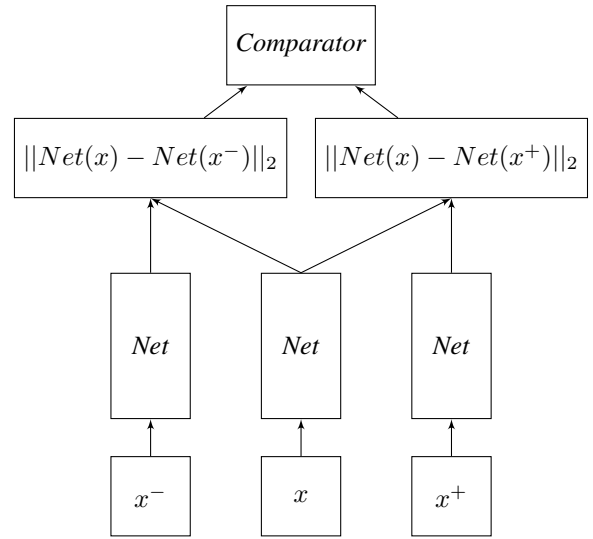

A Triplet Network is composed using three copies of the same neural network with shared parameters as shown in Fig.~\ref{fig:triplet_architecture}. The training examples consist of three samples, the anchor sample $x$, the positive sample $x^+$ and the negative sample $x^-$. The samples $x$ and $x^+$ are of the same class while $x^-$ is of a different class. The last layer of the neural network computes a representation $Net(x)$. The objective is to maximize the distance between inputs of different classes $|Net(x)-Net(x^-)|$ and minimize the distance of inputs belonging to the same class $|Net(x)-Net(x^+)|$. To achieve this, training uses the loss function: 
\begin{multline*}
    Loss(x,x^+,x^-)=max(|Net(x)-Net(x^+)|-\\|Net(x)-Net(x^-)|+\alpha,0)
\end{multline*}
where $\alpha$ is the margin between positive and negative pairs.

The triplet network can be trained by randomly sampling anchor data from the training set and augmenting them with one training sample with the same label as the anchor's and one sample with a different label, randomly chosen. However, this method leads to slow training and low performance as samples that result to $|Net(x)-Net(x^-)|>>|Net(x)-Net(x^+)|+\alpha$ do not provide useful information. The training can be improved by carefully mining the training data \cite{xuan2019improved}. For each training iteration, first, the anchor training data are randomly chosen. For each anchor, the hardest positive sample is chosen, meaning a sample from the same class as the anchor that is located the furthest away from the anchor. Then, the triplets are formed by mining all the hard negative samples, meaning the samples that satisfy $|Net(x)-Net(x^-)|<|Net(x)-Net(x^+)|$. 

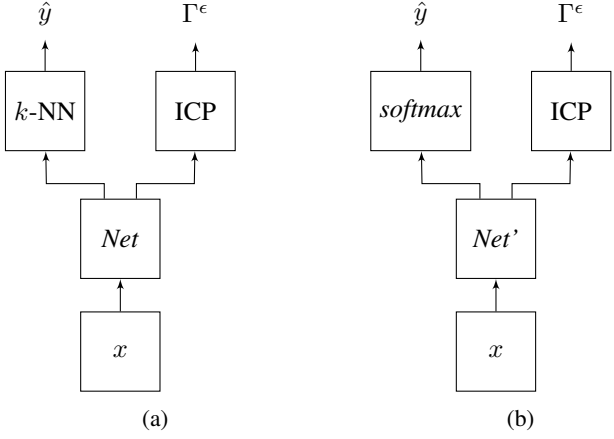
\begin{figure}[ht]
    \centering
    \begin{subfigure}[b]{4cm}
        \input{images/triplet_icp.tikz}
        \caption{}
        \label{fig:triplet_icp}
    \end{subfigure}
    \hfill
    \begin{subfigure}[b]{4cm}
        \input{images/dnn_icp.tikz}
        \caption{}
        \label{fig:dnn_icp}
    \end{subfigure}
    
    \caption{(a) LEC based on Triplet and (b) LEC based on DNN classifier}
    \label{fig:candidate_labels}
\end{figure}

A trained triplet network maps the raw data inputs to a lower dimensional space called \textit{embedding space}.
The distance between the representations $ Net(x) $ of the inputs is a useful measure of similarity and can be used for classification by training a $k$-NN classifier on the embedding space of the training data. This distance can also be used by the ICP framework as described in the next section.

%% file: images/triplet_architecture.tikz
\tikzstyle{block} = [draw, fill=none, rectangle, 
    minimum height=3em, minimum width=3em]
\tikzstyle{net_block} = [draw, fill=none, rectangle, 
    minimum height=5.5em, minimum width=3em]

\begin{tikzpicture}[auto, node distance=2cm,>=latex']
    \node [block] (x-) {$x^-$};
    \node [block, right of=x-, node distance=2cm] (x) {$x$};
    \node [block, right of=x,  node distance=2cm] (x+) {$x^+$};
    \node [net_block, above of=x-, node distance=2cm] (net-) {\textit{Net}};
    \node [net_block, above of=x, node distance=2cm] (net) {\textit{Net}};
    \node [net_block, above of=x+,  node distance=2cm] (net+) {\textit{Net}};
    \node [block, above of=net-, node distance=2.5cm] (diff-) {$||Net(x)-Net(x^-)||_2$};
    \node [block, above of=net+,  node distance=2.5cm] (diff+) {$||Net(x)-Net(x^+)||_2$};
    \node [block, above of=net, node distance=4cm] (comparator) {\textit{Comparator}};
    
    \draw[->] (x-) -- (net-);
    \draw[->] (x) -- (net);
    \draw[->] (x+) -- (net+);
    \draw[->] (net-) -- (diff-);
    \draw[->] ($(net.north east)!0.50!(net.north west)$) -- ($(diff-.south east)!0.50!(diff-.south west)$);
    \draw[->] ($(net.north east)!0.50!(net.north west)$) -- ($(diff+.south east)!0.50!(diff+.south west)$);
    \draw[->] (net+) -- (diff+);
    \draw[->] (diff-) -- (comparator);
    \draw[->] (diff+) -- (comparator);
\end{tikzpicture}
    

%% file: images/triplet_icp.tikz
\tikzstyle{block} = [draw, fill=none, rectangle, 
    minimum height=3em, minimum width=3em]
\tikzstyle{input} = [coordinate]
\tikzstyle{output} = [draw=none, fill=none, circle]
\tikzstyle{pinstyle} = [pin edge={to-,thin,black}]
\tikzstyle{branch}=[fill,shape=circle,minimum size=3pt,inner sep=0pt]

\begin{tikzpicture}[auto, node distance=1.5cm,>=latex']
    \node [block] (x) {$x$};
    \node[coordinate, left of=x, node distance=1cm] (x_left){};
    \node[coordinate, right of=x, node distance=1cm] (x_right){};
    \node [block, above of=x] (net) {\textit{Net}};
    \node [block, above of=x_left, node distance=3.2cm] (knn) {$k$-NN};
    \node [block, above of=x_right, node distance=3.2cm] (icp) {ICP};
    \node [output,above of=knn, node distance=1.3cm] (out) {$\hat{y}$};
    \node [output,above of=icp, node distance=1.3cm] (set) {$\Gamma^\epsilon$};
    
    \draw[->] (x) -- (net);
    \draw[->] ($(net.north east)!0.70!(net.north west)$) -- +(0,0.2) -- +(-0.77,0.2) -- (knn);
    \draw[->] ($(net.north east)!0.30!(net.north west)$) -- +(0,0.2) -- +(0.77,0.2) -- (icp);
    \draw[->] (knn) -- (out);
    \draw[->] (icp) -- (set);
    
\end{tikzpicture}

%% file: images/dnn_icp.tikz
\tikzstyle{block} = [draw, fill=none, rectangle, 
    minimum height=3em, minimum width=3em]
\tikzstyle{input} = [coordinate]
\tikzstyle{output} = [draw=none, fill=none, circle]
\tikzstyle{pinstyle} = [pin edge={to-,thin,black}]
\tikzstyle{branch}=[fill,shape=circle,minimum size=3pt,inner sep=0pt]

\begin{tikzpicture}[auto, node distance=1.5cm,>=latex']
    \node [block] (x) {$x$};
    \node[coordinate, left of=x, node distance=1cm] (x_left){};
    \node[coordinate, right of=x, node distance=1cm] (x_right){};
    \node [block, above of=x] (net) {\textit{Net'}};
    \node [block, above of=x_left, node distance=3.2cm] (softmax) {\textit{softmax}};
    \node [block, above of=x_right, node distance=3.2cm] (icp) {ICP};
    \node [output,above of=knn, node distance=1.3cm] (out) {$\hat{y}$};
    \node [output,above of=icp, node distance=1.3cm] (set) {$\Gamma^\epsilon$};
    
    \draw[->] (x) -- (net);
    \draw[->] ($(net.north east)!0.70!(net.north west)$) -- +(0,0.2) -- +(-0.77,0.2) -- (softmax);
    \draw[->] ($(net.north east)!0.30!(net.north west)$) -- +(0,0.2) -- +(0.77,0.2) -- (icp);
    \draw[->] (softmax) -- (out);
    \draw[->] (icp) -- (set);
    
\end{tikzpicture}

%% file: background.tex
\section{Triplet-Based ICP}
\label{sec:background}
In this section, we briefly explain the Inductive Conformal Prediction (ICP) approach based on the Triplet Network architecture. We consider a sequence of training examples, $z_1,\dots,z_l$ from $\bm{Z}$, where each $z_i$ is a pair $(x_i,y_i)$ with $x_i$ the feature vector and $y_i$ the corresponding label. We also consider a test input $x_{l+1}$ which we wish to classify. ICP assumes that all the examples $z_1,\dots,z_{l+1}$ are independent and identically distributed (IID) generated from the same but usually unknown probability distribution. 

For a chosen significance level $\epsilon\in [0,1]$, the objective is to generate a set of possible labels $\Gamma^\epsilon$ for the input $x_{l+1}$, such that the probability of the correct label $y_{l+1}\notin\Gamma^\epsilon$ does not exceed $\epsilon$. ICP is based on a \textit{nonconformity measure} (NCM) which is a dissimilarity metric between an example $z_{l+1}$ and the examples of the training set $z_1,\dots,z_l$. There are many different possible NCMs that can be used \cite{boursinos2020assurance},\cite{balasubramanian2014conformal},\cite{Vovk:2005:ALR:1062391},\cite{Shafer:2008:TCP:1390681.1390693},\cite{model_agnostic}. The proposed approach uses the Triplet Network to estimate similarities by encoding the inputs to an embedding space where the Euclidean distance between two samples is a direct measure of similarity. NCMs defined in the embedding space include (1) the \textit{k-Nearest Neighbors} ($k$-NN)~\cite{papernot2018deep}, (2) the \textit{one Nearest Neighbor} (1-NN)~\cite{Vovk:2005:ALR:1062391} and (3) the \textit{Nearest Centroid}~\cite{balasubramanian2014conformal}.

The $k$-NN NCM finds the $k$ most similar examples in the training data and counts how many of those are labeled different than the candidate label $y$ of a test input $x$. We denote $f:~X\rightarrow~V$ the mapping from the input space $X$ to the embedding space $V$ defined by the Triplet's last layer. After the training of the Triplet is complete, we compute and store the encodings $v_i=f(x_i)$ for the training data $ x_i $. Given a test example $x$ with encoding $v=f(x)$, we compute the $k$-nearest neighbors in $V$ and store their labels in a multi-set $\Omega$. The $k$-NN nonconformity of the test example $ x $ with a candidate label $y$ is defined as:
\begin{equation*}
\label{eq:nonconformity_multiple_neighbors}
\alpha(x,y)=|i\in\Omega:i\neq y|.
\end{equation*}

The 1-NN NCM requires to find the most similar example in the training set that has the same label as the candidate label $y$ of a test input $x$ as well as the most similar example that belong to any other class other than $y$. It is defined as: 
\begin{equation*}
\label{eq:nonconformity_1neighbor}
\alpha(x,y)=\dfrac{\min_{i=1,\dots,n:y_i=y}d(v,v_i)}{\min_{i=1,\dots,n:y_i\neq y}d(v,v_i)}
\end{equation*}
where $ v = f(x) $, $v_i=f(x_i)$, and $d$ is the euclidean distance metric in the $V$ space.

The Nearest Centroid NCM simplifies the task of computing individual training examples that are similar to a test example when there is a large amount of training data. We expect examples that belong to a particular class to be similar to each other so for each class $y_i$ we compute its centroid $\mu_{y_i}=\frac{\sum_{j=1}^{n_i}v_j^i}{n_i}$, where $v_j^i$ is the embedding representation of the  $ j^{th} $ training example from class $y_i$ and $n_i$ is the number of training examples in class $y_i$. The nonconformity function is defined as:
\begin{equation*}
\label{eq:nonconformity_nearest_centroid}
\alpha(x,y)=\dfrac{d(\mu_y,v)}{\min_{i=1,\dots,n:y_i\neq y}d(\mu_{y_i},v)}
\end{equation*}
where $ v = f(x) $. It should be noted that for computing the nearest centroid NCM, we need to store only the centroid for each class.

The NCM $a(x,y)$ is a measure of dissimilarity between a test input $x$ with candidate label $y$ and the training data $z_1\dots z_l$ as larger values would indicate higher "dissimilarity". However, this measure does not provide useful information by itself but it can be used by comparing it with NCM values computed using a \textit{calibration set} of known labeled data.  Consider the training set ($z_1\dots z_l$). This set is split into two parts, the proper training set ($z_1\dots z_m$) of size $m<l$ that will also be used for the training of the Triplet Network and the calibration set ($z_{m+1}\dots z_l$) of size $l-m$. The method first computes the NCMs $a(x_i,y_i)$, $i=m+1\dots l$ for the examples in the calibration set. Given a test example $x$ with an unknown label $y$, the method forms a set $|\Gamma^\epsilon|$ of possible labels $\tilde{y}$ so that $P(y\notin|\Gamma^\epsilon|)<\epsilon$. For all the candidate labels $\tilde{y}$, ICP is based on the empirical $p$-value defined as
\begin{equation*}
\label{eq:p_values_equation}
p_j(x)=\frac{|\{\alpha\in A:\alpha\geq\alpha(x,j)\}|}{|A|}.
\end{equation*}
that computes the fraction of nonconformity scores of the calibration data that are equal or larger than the nonconformity score of a test input. A candidate label is added to $\Gamma^\epsilon$ if $ p_j(x) > \epsilon $.

Depending on the desired significance level $\epsilon$, there may be more than one possible label. Although these multiple labels can provide useful information, we assume we do not have a way to process more than one possible label and we want to minimize these cases. The objective of the monitoring algorithm is after receiving each input to compute a valid prediction that ensures a predefined error rate (based on $\epsilon$) and limits the number of input examples for which a confident prediction cannot be made. The output of the monitor is defined as:
\begin{equation*}
    out =
        \begin{cases}
            0,& \text{if } |\Gamma^\epsilon|=0\\
            1,& \text{if } |\Gamma^\epsilon|=1\\
            \text{reject},& \text{if } |\Gamma^\epsilon|>1
        \end{cases}
\end{equation*}

Depending on the size of the set $\Gamma^\epsilon$, the monitor outputs $ out = 1 $ to indicate there is a single prediction with bounded confidence by $\epsilon$.  If the set $\Gamma^\epsilon$ is either empty or have multiple labels, the monitor raises alarms to indicate that no decision can be made. However, we distinguish between multiple and no predictions, because they may lead to different action in the system. An empty $\Gamma^\epsilon$ may be an indication that the input is out-of-distribution while multiple possible labels indicates that the accuracy of the underlying model is lower than the chosen $\epsilon$.

\begin{algorithm}[ht]
	\caption{\textbf{-- Monitoring Algorithm}.}
	\label{alg:approach}
	\begin{algorithmic}[1] 
		\Require training data $(X, Y)$, calibration data $(X^c, Y^c)$
		\Require trained neural network $f$ with $l$ layers
		\Require Nonconformity function $\alpha$
		\Require test input $z$
		\Require significance level threshold $\epsilon$
		\State // Compute the nonconformity scores for the calibration data offline
		\State $A=\{\alpha(x,y) : (x, y) \in (X^c, Y^c)\}$ \Comment{Calibration}
		\State // Generate prediction sets for each test data online
		\For{each label $j\in 1..n$}
		\State Compute the nonconformity score $\alpha(z,j)$
		\State $p_j(z) = \frac{\left| \{ \alpha \in A : \alpha \geq \alpha(z, j) \} \right|}{|A|}$ \Comment{empirical $p$-value}
		\If{$p_j(z)\geq\epsilon$}
		\State Add $j$ to the prediction set $\Gamma^\epsilon$
		\EndIf
		\EndFor
		\If{$|\Gamma^\epsilon|=0$}
		\State \Return 0
		\ElsIf{$|\Gamma^\epsilon|=1$}
		\State \Return 1
		\Else
		\State\Return Reject
		\EndIf
	\end{algorithmic}
\end{algorithm}

%% file: evaluation.tex
\section{Evaluation}
\label{sec:evaluation}
In this section, we evaluate how the triplet network improves the computation of significance levels using ICP. For the comparison, we use metrics that include how well inputs for different classes are clustered in the embedding represenation learned by the Triplet, the validity and efficiency of the produced prediction sets $|\Gamma^\epsilon|$, the execution time of the monitoring algorithm, and the required memory.

\subsection{Experimental Setup}
We apply the proposed method to the SCITOS-G5 wall following robot navigation dataset~\cite{Dua:2019}. The robot has the task of navigating around the room counter-clockwise in close proximity to the walls. It is equipped with 24 ultrasound sensors that are sampled at a rate of 9 samples per second. The possible actions are "Move-Forward", "Sharp-Right-Turn", "Slight-Left-Turn", and "Slight-Right-Turn". The SCITOS-G5 dataset contains 5456 raw values of the ultrasound sensor measurements as well as the decision it took in each sample. 10\% of the samples is used for testing. From the remaining 90\% of the data, 80\% is used for training and 20\% for calibration and/or validation.

The triplet network is formed using three identical DNNs with shared weights as shown in Fig. \ref{fig:triplet_architecture}. Since the inputs in the SCITOS-G5 dataset come from 24 sensors, we treat them as vectors and use a fully connected neural network. 
The parameters of the network architecture are shown in Table~\ref{tab:DNN_architecture}. For the baseline approach the DNN of Table~\ref{tab:DNN_architecture} is trained using the dedicated training data. The output of the hidden layer FC4, without taking into account the ReLu activation, 
is used to produce the representations for the baseline ICP. The FC5 is used as the baseline DNN classifier. The Triplet is trained using three copies of the same base DNN with hidden layers FC1-FC4. After training, only one of the three copies is used to generate the embeddings on the output of FC4 without any activation.
\input{tables/DNN_layers.tex}

All the experiments run in a desktop computer equipped with and Intel(R) Core(TM) i9-9900K CPU and 32 GB RAM and a Geforce RTX 2080 GPU with 8 GB memory.

\subsection{Triplet Performance}
We first look at how well the triplet clusters the data on a 16-dimensional space.
For comparison, we use the embedding space produced by the FC4 hidden layer of the DNN classifier.
A commonly used metric of the separation between classes is the \textit{Silhouette}~\cite{ROUSSEEUW198753}. For each sample, we first compute 
the mean distance between $i$ and all other data points in the same cluster in the embedding space
$$a(i)=\frac{1}{|C_i|-1}\sum_{j\in C_i, i\neq j}d(i,j)$$.

Then we compute the smallest mean distance from $i$ to all the data points in any other cluster
$$b(i)=\min_{k\notin i}\frac{1}{|C_k|}\sum_{j\in C_k}d(i,j)$$.

The silhouette value is defined as
$$s(i)=\frac{b(i)-a(i)}{\max\{a(i),b(i)\}}$$.
Each sample $i$ in the embedding space is assigned a silhouette value $-1\leq s(i)\leq 1$ depending on how close and how far it is to samples belonging to the same and different classes respectively. The closer $s(i)$ is to 1, the closer the sample is to samples of the same class and further from samples belonging to other classes. To compare the representations learned using the Triplet with the representations learned using the FC4 hidden layer of the fully connected DNN,
we compute the mean silhouette over the training data and the validation data separately. In Table \ref{tab:silhouettes}, we see that the representations learned by the Triplet form better-defined clusters.
\input{tables/silhouette_comparison.tex}

Another way to evaluate the performance of the triplet network is to combine it with a $k$-Nearest Neighbors classifier as shown in Figure~\ref{fig:triplet_icp}, and compute the classification accuracy. A test input is converted to an embedding representation using the triplet and then the $k$-NN classifier with $k$=15 is used to make a prediction. The baseline DNN uses a softmax activation (Figure~\ref{fig:dnn_icp}) for the classification. The Table~\ref{tab:accuracies} shows the accuracy of the triplet combined with a $k$-NN compared to the softmax classifier both using the same 
base neural network architecture shown in Table~\ref{tab:DNN_architecture}.
The triplet network results in a better performance than the DNN classifier with softmax.
\input{tables/triplet_classifier.tex}

\subsection{ICP and Assurance Monitoring}
The SCITOS-G5 robot is a safety-critical CPS for which the controller needs to take decisions with a well-calibrated and valid predefined significance level. 
As a baseline, we consider a NCM defined based on the penultimate layer of a DNN to generate a representations that allow real-time monitoring \cite{boursinos2020assurance}.
We evaluate the performance of the triplet when used with the nonconformity functions presented in \ref{sec:background}. In particular, we compare the calibration and validity of the predicted candidate label sets as well as we compute the bound $\epsilon$ that eliminates the sets $\Gamma^\epsilon$ with $|\Gamma^\epsilon|>1$, that is the sets with multiple predictions.

\begin{figure}[ht]
\centering
\includegraphics[width=0.9\linewidth]{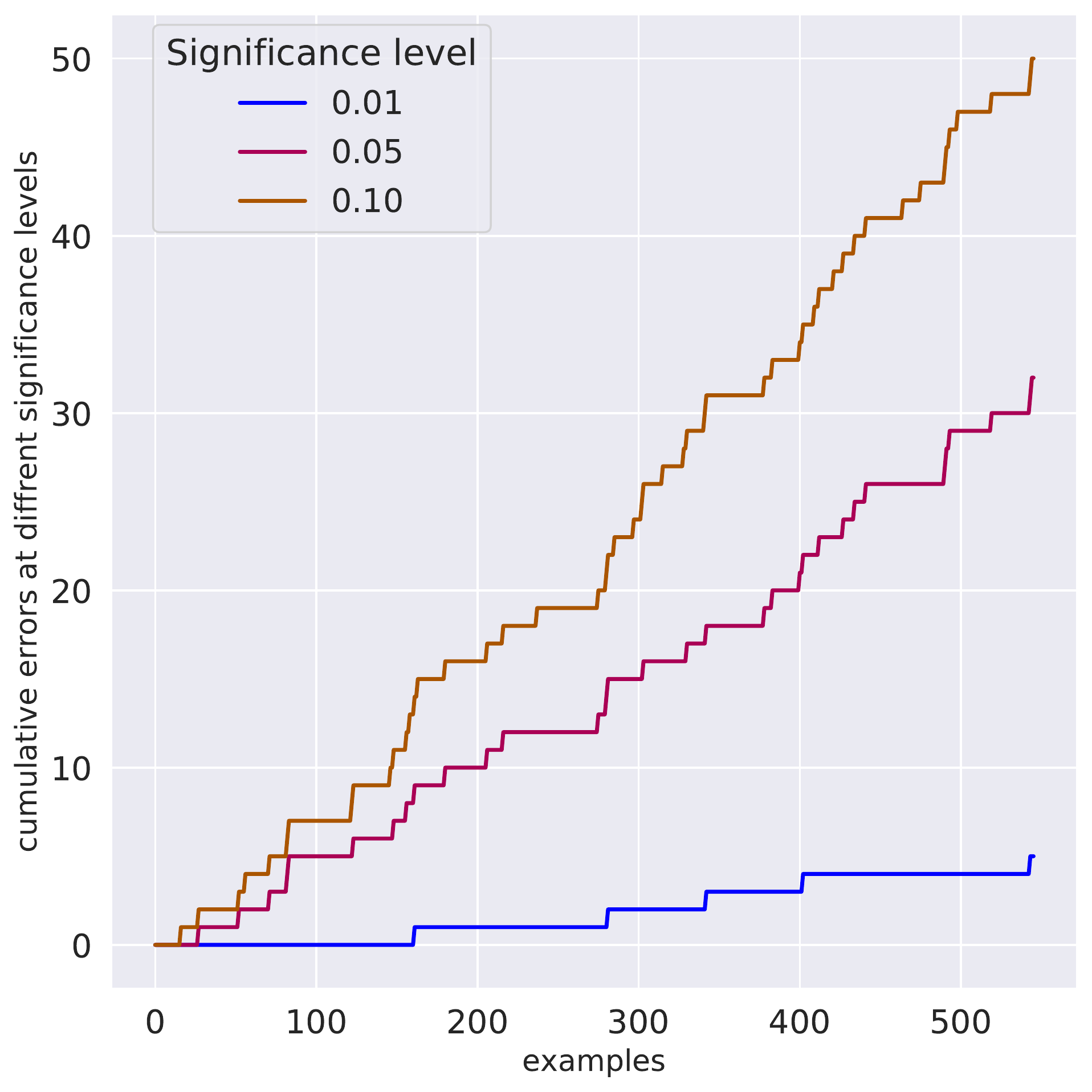}
\caption{Cumulative error curve}
\label{fig:errors_curve_plot}
\end{figure}

First, we verify that the error rates of the monitoring algorithm are bounded by the significance level $\epsilon$. We compute the percentage of incorrect predictions and we plot the cumulative error for different values of $\epsilon$. In Figure \ref{fig:errors_curve_plot}, we plot the cumulative error for three different values of $\epsilon$ for the SCITOS-G5 dataset using the Nearest Centroid nonconformity function. We see that the three different significance levels bound the cumulative error rate well. Similar behavior is observed using the other two nonconformity functions.

\begin{figure}[ht]
\centering
\includegraphics[width=0.9\linewidth]{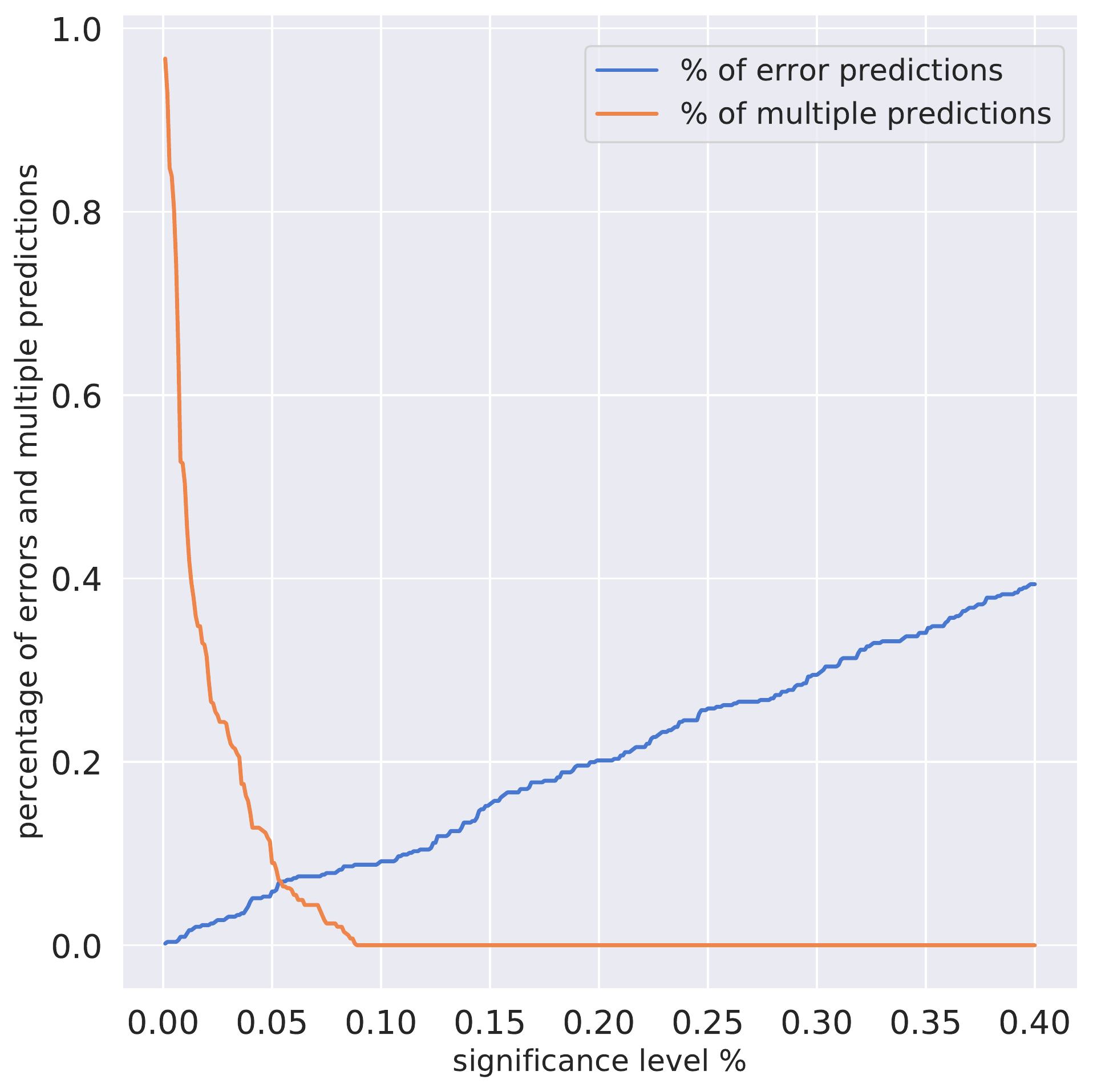}
\caption{Calibration and performance curve}
\label{fig:performance_curve_plot}
\end{figure}

In order to see how well-calibrated the confidence bounds are as well as how many sets with multiple candidate predictions are generated when $\epsilon\in[0.001,0.4]$, we plot the calibration and performance curves in Figure~\ref{fig:performance_curve_plot}. The number of multiple predictions decreases fast as $\epsilon$ increases. Further, the error rate is well-calibrated and increases linearly with $\epsilon$. 

\input{tables/epsilon_performance.tex}

Table~\ref{tab:epsilon_performance} reports the average execution time for each test input and the required memory using different nonconformity functions. The memory needed for the ICP application with each NC function is the memory required to store the DNN weights and the memory required to store the training data used by each NC function. In the $k$-NN and $1$-NN case, the encodings of the training data are stored in a $k-d$ tree~\cite{Bentley:1975:MBS:361002.361007} that is used to compute efficiently the nearest neighbors. In the $1$-NN case, it is required to find the nearest neighbor in the training data for each possible class which is computationally expensive resulting in larger execution time. The nearest centroid nonconformity function requires storing only the centroids for each class so the additional memory required is minimal. Both the triplet network architecture and the FC DNN we use as a baseline have the same network architecture and generate the same embedding size so the memory requirements are exactly the same between the two.

The evaluation results demonstrate that both the baseline FC DNN based ICP confidence monitor and the Triplet based ICP confidence monitor have well-calibrated error rates. The approach allows selecting the significance level to trade-off errors and alarms. When our objective is to minimize the number of alarms raised because of multiple candidate predictions, the Triplet with $k$-NN or Nearest Centroid NC functions satisfies it while keeping a higher significance level. When the significance level is selected, the Triplet based algorithm produces less sets of multiple candidate predictions. Finally, the memory requirements and the execution times for both approaches show that this monitoring system can be used for real-time CPS applications.

%% file: tables/DNN_layers.tex
\begin{table}[ht]
\centering
\begin{tabular}{|c|c|c|}
\hline
Layer & Size & Activation \\ \hline
FC 1  & 128  & ReLu       \\ \hline
FC 2  & 128  & ReLu       \\ \hline
FC 3  & 128  & ReLu       \\ \hline
FC 4  & 16   & ReLu       \\ \hline
FC 5  & 4    & Softmax    \\ \hline
\end{tabular}
\caption{The DNN architecture}
\label{tab:DNN_architecture}
\end{table}

%% file: tables/silhouette_comparison.tex
\begin{table}[ht]
\centering
\begin{tabular}{c|c|c|}
\cline{2-3}
                                         & Training Silhouette & Validation Silhouette \\ \hline
\multicolumn{1}{|c|}{Triplet Embeddings} & 0.52                & 0.46                  \\ \hline
\multicolumn{1}{|c|}{DNN Embeddings}     & 0.17                & 0.17                  \\ \hline
\end{tabular}
\caption{Clustering comparison using the silhouette coefficient}
\label{tab:silhouettes}
\end{table}

%% file: tables/triplet_classifier.tex
\begin{table}[ht]
\centering
\begin{tabular}{c|c|c|}
\cline{2-3}
                                     & Training Accuracy & Testing Accuracy \\ \hline
\multicolumn{1}{|c|}{Triplet + k-NN} & 95\%             & 91.2\%               \\ \hline
\multicolumn{1}{|c|}{DNN + Softmax}  & 91.93\%             & 88.49\%               \\ \hline
\end{tabular}
\caption{Comparison of the classification accuracy between a $k$-NN classifier on the triplet embeddings and a baseline DNN with softmax layer}
\label{tab:accuracies}
\end{table}

%% file: tables/epsilon_performance.tex
\begin{table*}[ht]
\centering
\begin{tabular}{c|c||c|c||c|c||c|c||c|c|}
\cline{3-10}
\multicolumn{2}{c|}{}                                             & \multicolumn{2}{c||}{Estimate $\epsilon$} & \multicolumn{2}{c||}{$\epsilon=0.01$} & \multicolumn{2}{c||}{$\epsilon=0.05$} & \multicolumn{2}{c|}{Runtime Requirements} \\ \hline
\multicolumn{1}{|c||}{Architecture}             & NC Functions     & $\epsilon$            & Errors           & Errors          & Multiples          & Errors          & Multiples          & Memory               & Time               \\ \hline\hline
\multicolumn{1}{|c||}{\multirow{3}{*}{Triplet}} & $k$-NN           & 0.089                 & 9.3\%            & 0\%             & 100\%              & 4\%             & 11.9\%             & 1.02 MB              & 1.4ms              \\ \cline{2-10} 
\multicolumn{1}{|c||}{}                         & 1-NN             & 0.087                 & 6.8\%            & 0.2\%           & 44.3\%             & 2.4\%           & 13.2\%             & 3.5 MB               & 2.8ms              \\ \cline{2-10} 
\multicolumn{1}{|c||}{}                         & Nearest Centroid & 0.092                 & 8.8\%            & 0.9\%           & 50.4\%             & 5.8\%           & 9\%                & 180 kB               & 1.1ms              \\ \hline\hline
\multicolumn{1}{|c||}{\multirow{3}{*}{FC DNN}}  & $k$-NN           & 0.095                 & 10.2\%           & 0\%             & 100\%              & 1.5\%           & 39.7\%             & 1.02 MB              & 1.6ms              \\ \cline{2-10} 
\multicolumn{1}{|c||}{}                         & 1-NN             & 0.08                  & 7.9\%            & 1.4\%           & 33.9\%             & 2.5\%           & 22.2\%             & 3.5 MB               & 3ms                \\ \cline{2-10} 
\multicolumn{1}{|c||}{}                         & Nearest Centroid & 0.201                 & 19.4\%           & 1.6\%           & 75.3\%             & 1.6\%           & 69\%               & 180 kB               & 1.4ms              \\ \hline 
\end{tabular}
\caption{Test results for different values of $\epsilon$ }
\label{tab:epsilon_performance}
\end{table*}

%% file: conclusion.tex
\section{Conclusion}
\label{conclusion}
DNN components are being used in Cyber-physical systems (CPS) to perform tasks like perception and control. 
However, their decision making process cannot be interpreted in a straightforward way and they cannot provide well-calibrated probabilities on the correctness of their predictions. 
This paper considers the problem of complementing the prediction of DNNs with a computation of confidence for real-time monitoring of CPS. We used the Inductive Conformal Prediction framework for classification tasks to produce sets of predictions with an error rate bounded by a chosen significance level. The evaluation results demonstrate that the addition of Triplet to the ICP framework produces well-calibrated probabilities as well as less prediction sets with more than one candidate decisions for a given significance level. The Triplet networks have been used on applications with higher dimensional input space such as images. Thus our future work is to scale our suggested assurance monitoring approach to be used on ICP applications that deal with images, like the camera outputs on self driving vehicles. Moreover, since there are many NC functions that can be used it is natural to see if we can combine the prediction sets produced by each of them on a test input using ensemble methods with voting to produce a narrower prediction set. 

%% file: aknowledgement.tex
\section*{Acknowledgment}
The material presented in this paper is based upon work supported by the Defense Advanced Research Projects Agency (DARPA) through contract number FA8750-18-C-0089 and the Air Force Office of Scientific Research (AFOSR) DDDAS through contract number FA9550-18-1-0126. The views and conclusions contained herein are those of the authors and should not be interpreted as necessarily representing the official policies or endorsements, either expressed or implied, of DARPA or AFOSR